\documentclass[conference]{IEEEtran}
\IEEEoverridecommandlockouts
\usepackage{cite}
\usepackage{amsmath,amssymb,amsfonts}
\usepackage{algorithm}
\usepackage{algorithmicx}
\usepackage{algpseudocode}
\usepackage{graphicx}
\usepackage{textcomp}
\usepackage{xcolor}
\usepackage{xspace}
\usepackage{soul}
\usepackage{kotex}
\usepackage{booktabs} % For formal tables
\usepackage{hyperref}
\definecolor{linkcolour}{rgb}{0,0.2,0.6}
\definecolor{xgreen}{rgb}{0.2,0.6,0.0}
\definecolor{xred}{rgb}{0.7,0.1,0.0}
\hypersetup{colorlinks,breaklinks,citecolor=red!50, urlcolor=linkcolour, linkcolor=xgreen}

\usepackage{xspace}
\newcommand{\BfPara}[1]{{\noindent\bf#1.}\xspace}

\usepackage{color}

\definecolor{pblue}{rgb}{0.13,0.13,1}
\definecolor{pgreen}{rgb}{0,0.5,0}
\definecolor{pred}{rgb}{0.9,0,0}
\definecolor{ppurple}{rgb}{0.5,0,0.5}

\usepackage{subfigure}
\usepackage{xcolor}
\usepackage{tikz}
\usepackage{listings}
\colorlet{punct}{red!60!black}
\definecolor{background}{HTML}{EEEEEE}
\definecolor{delim}{RGB}{20,105,176}
\colorlet{numb}{magenta!60!black}

\lstdefinelanguage{json}{
    basicstyle=\normalfont\ttfamily,
    numbers=left,
    numberstyle=\small,
    stepnumber=1,
    numbersep=12pt,
    showstringspaces=false,
    breaklines=true,
    frame=lines,
    backgroundcolor=\color{background},
    literate=
     *{0}{{{\color{numb}0}}}{1}
      {1}{{{\color{numb}1}}}{1}
      {2}{{{\color{numb}2}}}{1}
      {3}{{{\color{numb}3}}}{1}
      {4}{{{\color{numb}4}}}{1}
      {5}{{{\color{numb}5}}}{1}
      {6}{{{\color{numb}6}}}{1}
      {7}{{{\color{numb}7}}}{1}
      {8}{{{\color{numb}8}}}{1}
      {9}{{{\color{numb}9}}}{1}
      {:}{{{\color{punct}{:}}}}{1}
      {,}{{{\color{punct}{,}}}}{1}
      {\{}{{{\color{delim}{\{}}}}{1}
      {\}}{{{\color{delim}{\}}}}}{1}
      {[}{{{\color{delim}{[}}}}{1}
      {]}{{{\color{delim}{]}}}}{1},
}

\def\BibTeX{{\rm B\kern-.05em{\sc i\kern-.025em b}\kern-.08em
    T\kern-.1667em\lower.7ex\hbox{E}\kern-.125emX}}
\begin{document}

\title{Introduction to Quantum Reinforcement Learning: Theory and PennyLane-based Implementation}

\author{\IEEEauthorblockN{$^{\circ}$Yunseok Kwak, $^{\circ}$Won Joon Yun, $^{\dag}$Soyi Jung, $^{\circ}$Jong-Kook Kim, and $^{\circ}$Joongheon Kim}
\IEEEauthorblockA{$^{\circ}$School of Electrical Engineering, Korea University, Seoul, Republic of Korea 
\\
$^{\dag}$School of Software, Hallym University, Chungcheon, Republic of Korea 
\\
E-mails: \texttt{rhkrdbstjr0@korea.ac.kr}, 
\texttt{ywjoon95@korea.ac.kr}, 
\texttt{jungsoyi@korea.ac.kr},\\
\texttt{jongkook@korea.ac.kr},
\texttt{joongheon@korea.ac.kr}
}
}
\maketitle

\begin{abstract}
The emergence of quantum computing enables for researchers to apply quantum circuit on many existing studies. Utilizing quantum circuit and quantum differential programming, many research are conducted such as \textit{Quantum Machine Learning} (QML). In particular, quantum reinforcement learning is a good field to test the possibility of quantum machine learning, and a lot of research is being done. This work will introduce the concept of quantum reinforcement learning using a variational quantum circuit, and confirm its possibility through implementation and experimentation. We will first present the background knowledge and working principle of quantum reinforcement learning, and then guide the implementation method using the PennyLane library. We will also discuss the power and possibility of quantum reinforcement learning from the experimental results obtained through this work.

\end{abstract}

\section{Introduction}
Since deep reinforcement learning has opened a new chapter in reinforcement learning by leveraging the power of artificial neural networks, there have been achievements such as surpassing human limits in complex games such as chess and Go, and it has already become an unavoidable flow of artificial intelligence (AI) research.

On the other hand, advanced quantum computing technology has already reached a level close to the implementation of quantum computational gain predicted through many algorithm studies~\cite{apwcs21kim,ictc19choi,icoin20choi}. In addition, the advent of a \textit{variational quantum circuit} (VQC) that mimics the principles and functions of artificial neural networks has made it possible to apply these quantum calculations to existing machine learning algorithms.
This has established itself as a major trend in quantum machine learning research, and many studies using it are being actively conducted. In this context, many studies are conducted using VQC as a \textit{Quantum Neural Network} (QNN)~\cite{arxiv01,ictc20oh,icoin21oh,icoin21choi}, including variational classifier, image preprocessor, federated learning, reinforcement learning, etc. Among them, in this paper, we introduce and discuss quantum reinforcement learning, a reinforcement learning model that replaces the artificial neural network of a \textit{deep Q network} (DQN) with a VQC. 

Since QPU has not been commercialized yet, there is nothing better in speed than the existing machine learning framework that utilizes NPU. However, as development libraries such as Tensorflow-quantum~\cite{broughton2020tensorflow}, Qiskit~\cite{Qiskit}, and PennyLane~\cite{bergholm2020pennylane} for future quantum computing environments and quantum computing clouds such as IBMQ, IonQ, and Amazon Braket are provided to developers, various studies on QML are in progress. Particularly, PennyLane~\cite{bergholm2020pennylane}, is a suitable library for starting quantum machine learning research because it provides a simulator that allows users to easily implement quantum circuits by using a CPU to perform QPU operations. Therefore, we aim to increase access to quantum reinforcement learning and facilitate subsequent research by briefly introducing the implementation process through PennyLane. In addition, we would like to discuss the impacts and potentials of quantum computing in reinforcement learning through experimentation and evaluation of quantum reinforcement learning models in the CartPole environment provided by OpenAI.

\section{Backgrounds}
\subsection{Reinforcement Learning}
Reinforcement learning is mathematically modeled with \textit{Markov Decision Process} (MDP) as a tuple $(\mathcal{S}, \mathcal{A}, P, R, T)$, where $\mathcal{S}$ is a finite set of state information, and $\mathcal{A}$ is a finite set of action information. The function $P : \mathcal{S} \times \mathcal{A} \to P(\mathcal{S})$ is a transition probability function, with $P(s'\mid s, a)$ being the probability of transitioning into state $s'$ if an agent starts executing action $a$ in state $s$. The function $R : \mathcal{S} \times \mathcal{A} \times \mathcal{S} \to \mathbb{R}$ denotes the reward function, with $R_{t} = R(s_t, a_t, s_{t+1})$. The MDP has a finite time horizon $T$, and solving an MDP means finding a optimal policy $\pi_\theta^*  \in \Pi : \mathcal{S} \times \mathcal{A} \to \left[ 0, 1 \right]$, where $\pi_\theta$ is neural network-based policy with parameter $\theta$; Observing $s$, $\pi_\theta$ determines agent's action $a \in \mathcal{A}$ to maximize the cumulative rewards received during the finite time $T$. 

When the environment transitions and the policy are stochastic, the probability of a $T$-step trajectory is defined as $P(\tau \mid \pi_{\theta}) = \rho(s_0)\prod_{t=0}^{T-1}P(s_{t+1}\mid s_t,a_t)\pi_{\theta}(a_t \mid s_t)$
where $\rho$ is the initial state distribution. Then, the expected return $\mathcal{J}(\pi_{\theta})$ is defined as $\mathcal{J}(\pi_{\theta}) = \int_{\tau}P(\tau \mid \pi_{\theta})R(\tau) = \mathbb{E}_{\tau \sim \pi_{\theta}}\left[R(\tau)\right]$
where the trajectory $\tau$ is a sequence of states and actions in the environment. 
The objective of reinforcement learning is to learn a policy that maximizes the expected return $\mathcal{J}(\pi_{\theta})$ when the agent acts according to the policy $\pi_{\theta}$. Therefore, the optimization objective is expressed by
\begin{equation}
    \label{eq:obj_rl}
\pi_{\theta}^* = \arg\max_{\theta} \mathcal{J}(\pi_{\theta})
\end{equation}
with $\pi_{\theta}^*$ being the optimal policy. 

\BfPara{Deep \textit{Q}-Network (DQN)\cite{mnih2013playing}}
One of the conventional method for solving MDP is \textit{Q}-Learning. \textit{Q}-Learning utilizes \textit{Q}-table to find optimal policy. However, \textit{Q}-Learning has limitation that it obtains optimal policy when the state dimension is small. Inspired to \textit{Q}-Learning, deep \textit{Q}-network (DQN) which is a model-free reinforcement learning, is proposed to learn the optimal policy with a high-dimensional state space.
Experience replay $\mathcal{D}$ and target network are two key features used for training deep neural network with stabilization. 
Experiences $e_t = (s_{t}, a_{t}, R_{t+1}, s_{t+1})$ of the agent are stored in the experience buffer $\mathcal{D} = (e_1, e_2, \dots, e_T)$, and are periodically resampled to train the $\textit{Q}$-networks. 
Sampled experience is used to update the parameters $\theta_i$ of the policy with the loss function at the $i$-th  training iteration where the loss function is defined as
\begin{multline}
    \label{eq:dqn}
    L(\theta_i) = \mathbb{E}\left[(R_{t+1} + \right.\\ \left.\gamma \max_{a'}Q(s_{t+1}, a' ; \theta_{i}^{-}) - Q(s_t, a_t ; \theta_{i}))^2\right]
\end{multline}
where $\theta^{-}_i$ are the target network parameters. The target network parameters $\theta^{-}_i$ are updated using the \textit{Q}-network parameters $\theta$ in every predefined step. The stochastic gradient descent method is used to optimize the loss function.

\BfPara{Proximal Policy Optimization (PPO)\cite{schulman2017proximal}}
PPO is one of the breakthroughs of DRL algorithms for improving the training stability by ensuring that $\pi_\theta$ updates at every iteration are small by clipping the probability ratio $r_{\pi}(\theta) = \pi_\theta(a \mid s) / \pi_{\theta_{old}}(a \mid s)$, where $\theta_{old}$ is that of previous updated parameters of policy. 
(Schulman et al., 2017) proposed a surrogate function that has objective that prevents the new policy from straying away from the old one is used to train the policy $\pi_\theta$. The clipped objective function is as follows:
\begin{equation}
    \label{eq:ppo_surrogate}
    L^{\text{CLIP}}_t(\theta) = \min(r_t(\theta)A_t, \textit{clip}(r_t(\theta),1-\epsilon,1+\epsilon)A_t),
\end{equation}where $A_t$ is the estimated advantage function under hyperparameter $\epsilon < 1$, which means how far away the new policy is allowed to update from the old policy. PPO uses the stochastic gradient descent to maximize the objective \eqref{eq:ppo_surrogate}.

\subsection{Quantum Computing}
Quantum computers use a qubit as the basic unit of computation, which represent a quantum superposition state between two basis state $|0\rangle$ and $|1\rangle$. It is controlled by unitary gates in a quantum circuit to perform various quantum operations. It can be represented as a normalized two-dimensional complex vector as: 
\begin{equation}
    \label{eq:qubit}    
|\psi\rangle = \alpha|0\rangle + \beta|1\rangle, ~\mathrm{where}~\|\alpha\|_2^2 + \|\beta\|_2^2 = 1,
\end{equation}
and there also is a geometrical representation of a qubit space, using polar coordinates $\theta$ and $\phi$:
\begin{equation}
    \label{eq:bloch}    
|\psi\rangle = \cos(\theta/2)|0\rangle + e^{i\phi}\sin(\theta/2)|1\rangle ,
\end{equation}
where $ 0\leq\theta\leq\pi$ and $ 0\leq\phi\leq\pi$.
Qubit state is mapped into the surface of 3-dimensional unit sphere, which is called \textit{Bloch sphere}. Quantum gate is a unitary operator transforming a qubit state into another qubit state, which can be represented as a $2\times2$ matrix with complex entries. There are some important quantum gates, Pauli-$X$, Pauli-$Y$, and Pauli-$Z$, rotating by $\pi$ around their corresponding axes in Bloch sphere. The rotation operator gates $R_x(\theta), R_y(\theta)$, and $R_z(\theta)$ rotate by $\theta$ instead of $\pi$ in Pauli-$X$, Pauli-$Y$, and Pauli-$Z$ gates, and it is known that any single-qubit unitary gate in $SU(2)$ can be written as 
a product of three rotation operators of each axis. In addition, there are quantum gates which operate on multiple qubits, called controlled rotation gates. They act on a qubit according to the signal of several control qubits, which generates quantum entanglement between multiple qubits. Among them, Controlled X(or CNOT) gate is one of the most used control gates, changing the sign of the second qubit if the first qubit is $|1\rangle$. These gates allow quantum algorithms to work using their features on a quantum circuit that will be introduced later.
\subsection{Variational Quantum Circuit}
The variational quantum circuit (or parameterized quantum circuit) is a quantum circuit using learnable parameters to perform various numerical tasks, such as optimization, approximation, and classification. Operation of general VQC model can be divided into 4 steps. First one is state preperation step, the input information is encoded into corresponding qubit states, which can be treated in the quantum circuit. Next step is variational step, entangling qubit states by controlled gates and rotating qubits by parameterized rotation gates. This process can be repeated in a multi-layer manner with more parameters, which possibly enhance the performance of the circuit. In the third step, processed qubit states are measured and decoded to the form of appropriate output information. Last step is conducted outside the circuit. The quantum circuit parameters are updated in the direction of optimizing the objective function of the algorithm by a classical CPU algorithm, like Adam optimizer. Then the circuit updated with the new parameters performs the calculation again from the beginning. This circuit is known to be able to approximate any continuous function like classical neural network\cite{biamonte2021universal}, so VQC is often called \textit{Quantum Neural Network} (QNN)\cite{wiebe2014quantum}. It has been widely applied in quantum machine learning researches.

\begin{algorithm*}[t]
\begin{algorithmic}\label{alg1}
\State Initialize replay memory $\mathcal{D}$ to capacity $N$
\State Initialize action-value function quantum circuit $Q$ with random parameters $\theta$
\State Initialize state value function $V(s;\phi)$
%\State Require preprocessor $h(s)$ that maps histories to fixed-length representations.
\For{episode $=1,2,\ldots,M$} 
\State Initialise state $s_1$ and encode into the quantum state
\State \textbf{\# 1. Inference Process \#}
    \For {$t=1,2,\ldots,T$}
    	\State With probability $\epsilon$ select a random action $a_t$
    	\State otherwise select $a_t = \max_{a} Q^*(s_t, a; \theta)$ from the output of the quantum circuit
    	\State Execute action $a_t$ in emulator and observe reward $r_t$ and next state $s_{t+1}$
    	\State Store transition $\left(s_t,a_t,R_t,s_{t+1}\right)$ in $\mathcal{D}$
    \EndFor
    \State \textbf{\# 2. Training Process \#}
    \For{$i=1,...,K_{epoch} $}
    	\State Sample random mini-batch of transitions $\left(s_j,a_j,R_j,s_{j+1}\right)$ from $\mathcal{D}$
    	\State Calculate temporal difference target,
    	$y_j =
        \left\{
        \begin{array}{l l}
          R_j  \quad & \text{for terminal } s_{j+1}\\
          R_j + \gamma \max_{a'} Q(s_{j+1}, a'; \theta) \quad & \text{for non-terminal } s_{j+1}
        \end{array} \right.$
        \State Calculate temporal difference, $\delta_j = y_j - V(s_j)$
    	\State Calculate estimated advantage function, $\hat{A}_{j} = \delta_{j} + (\gamma \lambda)\delta_{j+1}+...+ (\gamma \lambda)^{J-j+1}\delta_{J-1}$
    	\State Calculate ratio, $r_j= \frac{\pi_\theta(a_j|s_j)}{\pi_{\theta_{OLD}}(a_j|s_j)}$
	    \State Calculate surrogate actor loss function using \eqref{eq:ppo_surrogate}
	    \State Calculate critic loss function, $|V(s)-y_j|$.
	    \State Calculate gradient and update actor and critic parameters
	\EndFor
\EndFor
\end{algorithmic}
\caption{Variational Quantum Deep Q Learning with PPO}
\end{algorithm*}

\section{Quantum Reinforcement Learning}
\subsection{Variational Quantum Policy Circuit}
\begin{figure}[t!] \centering \includegraphics[width=1\columnwidth]{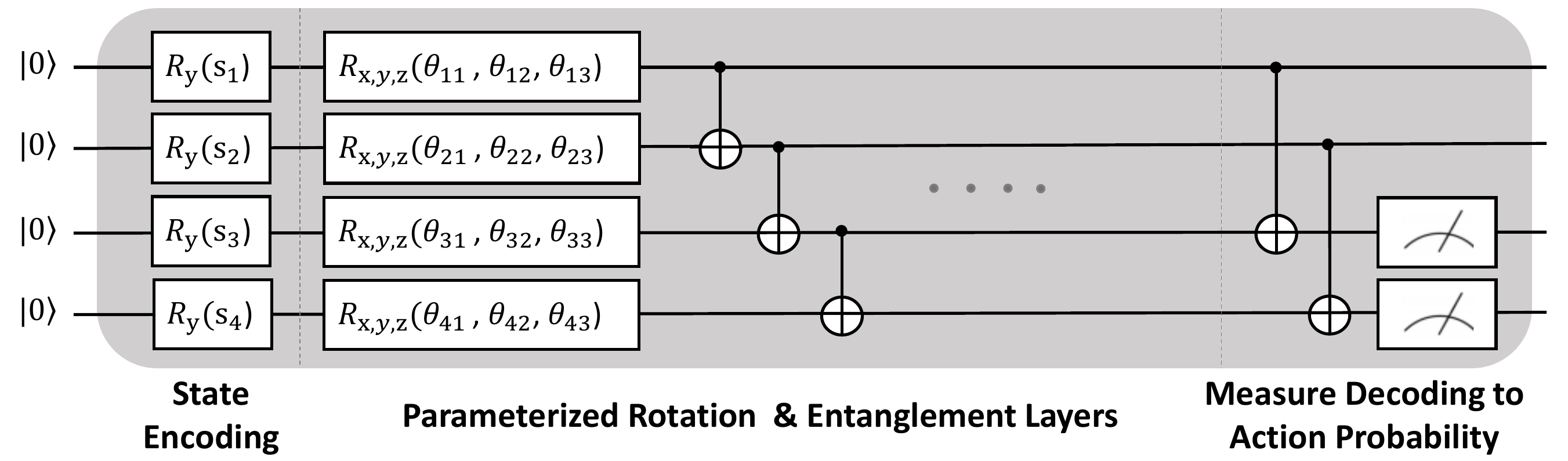} \caption{A policy-VQC for deep-Q learning with parameter $\theta$. } \label{fig:VQC} \end{figure} 

\begin{figure}[t!]
    \centering
    \includegraphics[width=1\columnwidth]{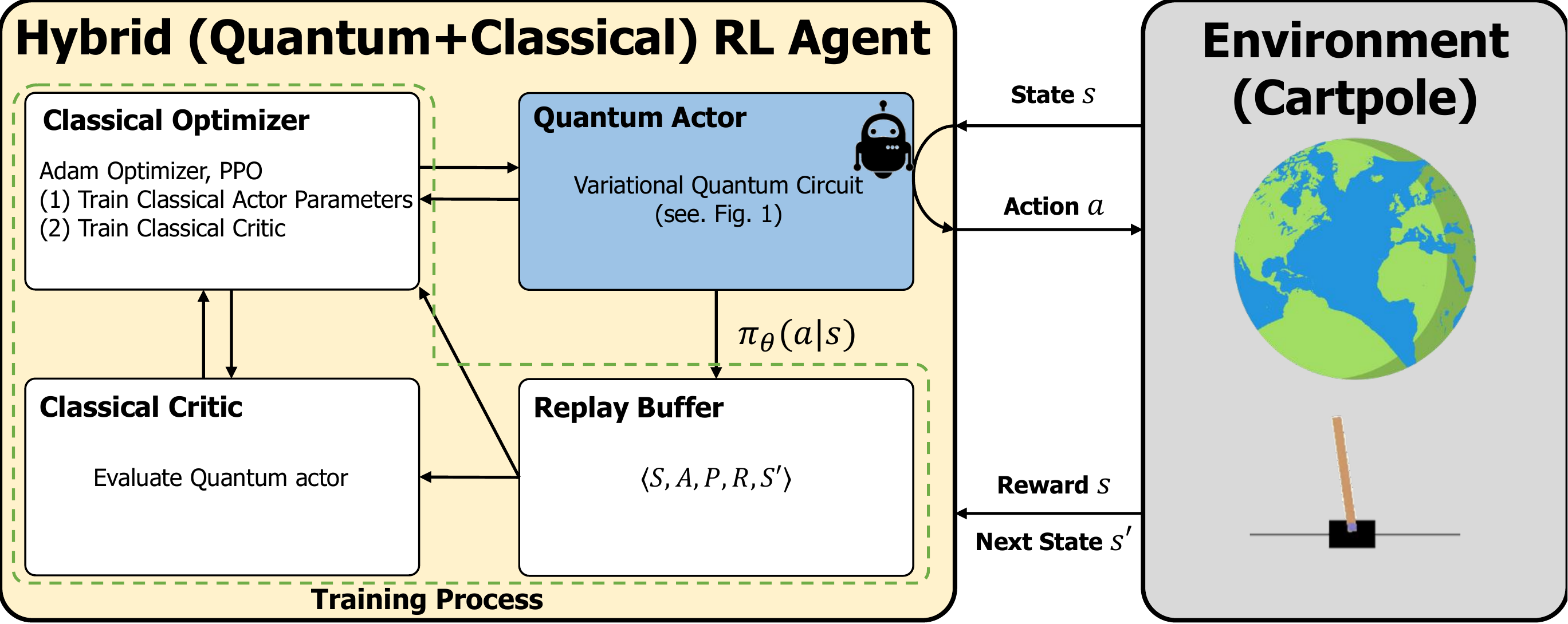}
    \caption{Quantum Reinforcement Learning System}
    \label{fig:Env}
\end{figure}
In recent studies of quantum reinforcement learning \cite{chen2020variational,jerbi2021variational}, VQC substitutes the policy training DNN of existing DRL. At each episode, agent with given state information determines its action from policy-VQC and parameters are updated with a classical CPU algorithm like Adam Optimizer. This paper approaches similarly, by using a VQC depicted in Fig.~\ref{fig:VQC}. \\
The quantum circuit in Fig.~\ref{fig:VQC} is a prototype of policy-VQC, which consists of the basic structure of VQC. State encoding part of the circuit includes $R_y$ gates parameterized by normalized state input $s$, having their values between $-\pi$ and $\pi$. 
Variational part in the center consists of entangling $CX$ Gates and $R_x$, $R_y$, $R_z$ gates parameterized with free parameter $\theta$. This part is called a layer, and several can be repeatedly stacked in a circuit. After that, measured output of the circuit is decoded into the action space, yielding the action probabilities. Then the obtained $\pi_\theta$ is evaluated and updated in a classical computer.

\subsection{Quantum Reinforcement Learning Systems}
The quantum reinforcement learning system in this paper is as described in Fig.~\ref{fig:Env}. In the beginning of an episode, quantum-classical hybrid agent receives a state information from the environment, and determine its action by $\theta_\pi$ made from the VQC. Then the policy of the agent is evaluated and updated by PPO algorithm, which is introduced before and described in Algorithm~\ref{alg1}. 
The PPO algorithm are effectively the same as in the previous study\cite{schulman2017proximal}. The replay buffer functions in the same way as in traditional approaches, keeping track of the $\langle s, a, R, s^\prime \rangle$ tuples. One does not have to fundamentally or drastically change an algorithm in order to apply the power of VQCs to it. The algorithm presented in Algorithm~\ref{alg1}.

\begin{figure*}[t]
    \centering
%\begin{minipage}{1.5\columnwidth}
\begin{lstlisting}[backgroundcolor = \color{gray!5}, framexleftmargin = 0em, aboveskip=0pt, belowskip = 0pt, frame = t, framesep = 0cm, framerule=0pt]
#Parameterized Rotation & Entanglement Layers
def layer(W):
    for i in range(n_qubit):
      qml.RX(W[i,0], wires=i)
      qml.RY(W[i,1], wires=i)
      qml.RZ(W[i,2], wires=i)
      
#Classical Critic
class V(nn.Module):
    def __init__(self):
        super(V, self).__init__()
        self.fc1   = nn.Linear(4,256)
        self.fc_v  = nn.Linear(256,1)
    def forward(self,x):
        x = F.relu(self.fc1(x))
        v = self.fc_v(x)
        return v
        
#Variational Quantum Policy Circuit (Actor)
@qml.qnode(dev, interface='torch')
def circuit(W,s): 
    # W: Layer Variable Parameters, s: State Variable
    
    # Input Encoding
    for i in range(n_qubit):
        qml.RY(np.pi*s[i],wires=i)
        
    #Variational Quantum Circuit
    layer(W[0])
    for i in range(n_qubit-1):
        qml.CNOT(wires=[i,i+1])
   layer(W[1])
    for i in range(n_qubit-1):
        qml.CNOT(wires=[i,i+1])
    layer(W[2])
    for i in range(n_qubit-1):
        qml.CNOT(wires=[i,i+1])
    layer(W[3])
    qml.CNOT(wires=[0,2])
    qml.CNOT(wires=[1,3])
    return [qml.expval(qml.PauliY(ind)) for ind in range(2,4)]
    
#Declare Quantum Circuit and Parameters
W   = Variable(torch.DoubleTensor(np.random.rand(4,4,3)),requires_grad=True)
v = V()
circuit_pi = circuit
optimizer1 = optim.Adam([W], lr=1e-3)
optimizer2 = optim.Adam(v.parameters(), lr=1e-5)
  \end{lstlisting}
%\end{minipage}
\caption{Variational Quantum Policy Circuit with PennyLane}
    \label{fig:fig1}
\end{figure*}

\section{The Implementation of QRL}
\subsection{Implementation Guidelines}
As quantum computing and quantum machine learning research is actively progressing, many development libraries for researchers have emerged, such as TensorFlow-quantum, Qiskit, and PennyLane. Among them, PennyLane was created to support quantum machine learning research, allowing anyone to easily test the performance of quantum circuits through quantum simulators. The quantum simulator supported by PennyLane allows the CPU to imitate the operation of QPU, and especially supports the use of parameters in the form of PyTorch tensor and gradient operation~\cite{NEURIPS2019_9015}. Thanks to these features, PennyLane makes it easy for anyone who has previously done machine learning research using PyTorch to start researching quantum machine learning. Based on this background, in this paper, a quantum reinforcement learning model was implemented using PennyLane and PyTorch as shown in Fig.~\ref{fig:fig1}.

\subsection{The CartPole Environment}
Cartpole, the implementation environment in this paper, is a test environment for reinforcement learning provided by OpenAI~\cite{1606.01540}. This is a game where the agent moves the cart back and forth to avoid dropping the stick on the cart and the longer one holds the stick, the greater the reward. At every moment, the player observes the cart's position, velocity, and the angle and angular velocity of the rod to determine which direction to accelerate accordingly. The VQC in Fig.~\ref{fig:VQC} using 4 qubits is suitable for policy making in this environment. Each of the four pieces of information provided by the environment is normalized and fed into the circuit as values between $-\pi$ and $\pi$, and the two measures are decoded into the probability values of taking two actions via the softmax function. This process continues until the agent can take the optimal action on the given state information by optimizing the given parameters in the reinforcement learning algorithm. The experimental result is showed later in this paper.
\subsection{Experimental Setup}
Our experiment is conducted with the software packages, PyTorch for speed and convenience of tensor operation and and PennyLane for quantum circuit simulation. The quantum simulator provided by Pennylane is very convenient to use, but it is difficult to use many qubits because of its slow computational speed. Therefore, the CartPole environment was used as a simple environment that can be operated with a circuit of small qubits. We used classical parameter optimizer as Adam optimizer with learning rate 0.001 for quantum policy and 0.00001 for classical critic. Other hyperparameter settings are  $\gamma$ = 0.98, $\lambda$ = 0.95, $\epsilon$ = 0.01. The baseline model is a random version of this model, using random parameters in every time step without optimization.
\subsection{Experimental Results}
 \begin{figure}[t!]
 \includegraphics[width=1\columnwidth]{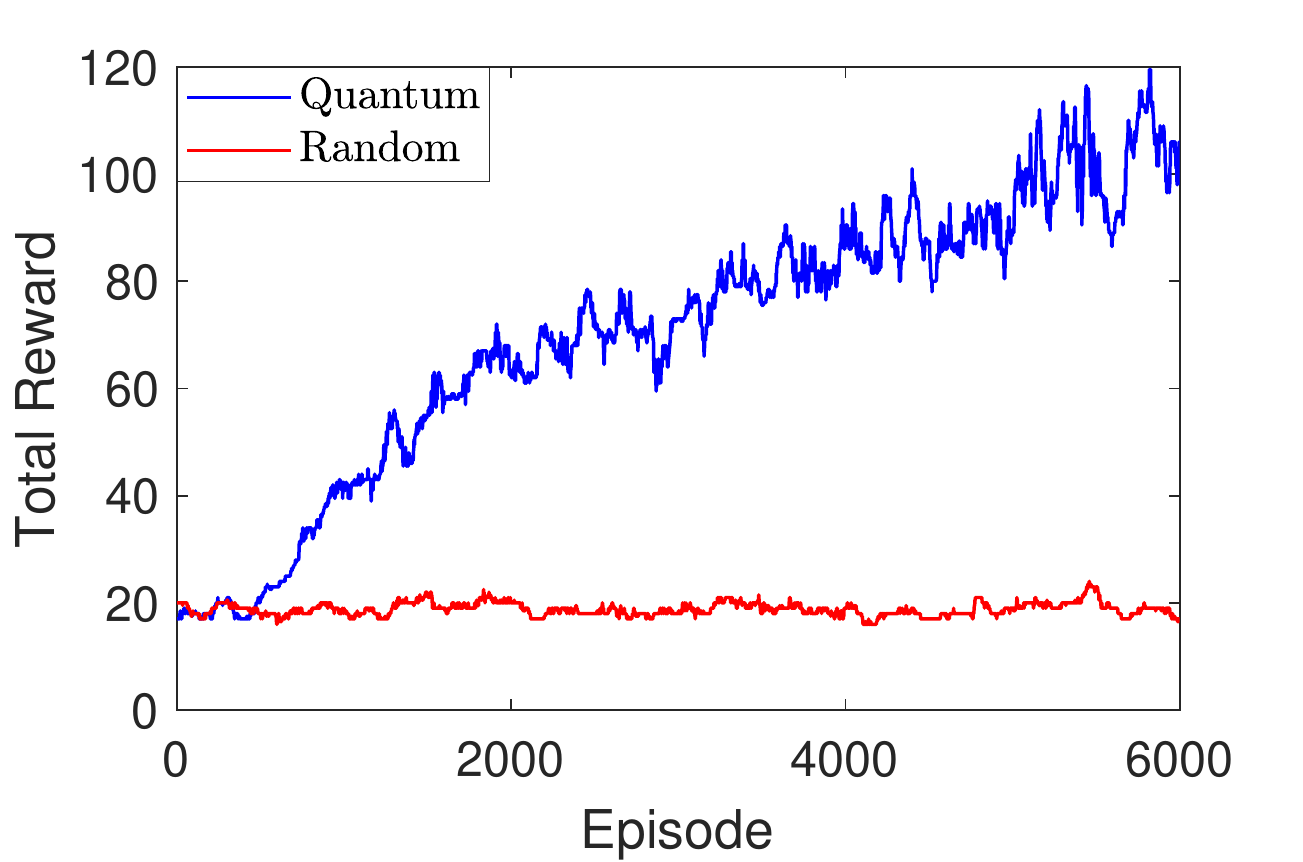}
 \caption{Comparison of Total Reward on Average in Environment (CartPole-v0)}
 \label{fig:exp_result}
 \end{figure}
%\cite{ton201608kim}
Fig.~\ref{fig:exp_result} shows the performance of the proposed quantum reinforcement learning model. Comparing with random actions, one can see that the model is learning to find the optimal action. Also, it can be seen that the deviation of rewards during the learning process is extremely high. This is due to uncertainty within quantum systems, although the impact of reinforcement learning algorithms facilitating exploration cannot be ignored either. This uncertainty simultaneously implies the possibilities and limitations of quantum reinforcement learning. This allows effective policy exploration with relatively few tens of parameters, but makes it difficult to maintain the good results once reached. Leveraging these characteristics is an important challenge for quantum reinforcement learning.
\section{Conclusions and Future Work}\label{sec:sec5}
Through this work, we implemented and tested a quantum reinforcement learning model in the CartPole environment based on PPO, one of the latest deep reinforcement learning techniques, and demonstrated implementation guidelines in the PennyLane library. We discussed the principle and potential of the reinforcement learning using a variational quantum circuit. Furthermore, this work aims to enable researchers who are new to quantum reinforcement learning to start research with interest. Although the performance of quantum reinforcement learning cannot be said to be better than that of the existing method, it is expected that many follow-up studies will yield results that exceed the limitations of existing reinforcement learning.

\section*{Acknowledgment}
This work was supported by the National Research Foundation of Korea (2019M3E4A1080391).
Joongheon Kim is a corresponding author of this paper.


\begin{thebibliography}{1}

\bibitem{apwcs21kim}
J.~Kim, Y.~Kwak, S.~Jung, and J.-H. Kim, ``Quantum scheduling for
  millimeter-wave observation satellite constellation,'' in \emph{Proceedings
  of the IEEE VTS Asia Pacific Wireless Communications Symposium (APWCS)},
  2021, pp. 1--1.

\bibitem{ictc19choi}
J.~Choi and J.~Kim, ``A tutorial on quantum approximate optimization algorithm
  {(QAOA)}: Fundamentals and applications,'' in \emph{Proceedings of the IEEE
  International Conference on Information and Communication Technology
  Convergence (ICTC)}, 2019, pp. 138--142.

\bibitem{icoin20choi}
J.~Choi, S.~Oh, and J.~Kim, ``The useful quantum computing techniques for
  artificial intelligence engineers,'' in \emph{Proceedings of the IEEE
  International Conference on Information Networking (ICOIN)}, 2020, pp. 1--3.

\bibitem{arxiv01}
Y.~Kwak, W.~J. Yun, S.~Jung, and J.~Kim, ``Quantum neural networks: Concepts,
  applications, and challenges,'' \emph{CoRR}, vol. abs/2108.01468, 2021.

\bibitem{ictc20oh}
S.~Oh, J.~Choi, and J.~Kim, ``A tutorial on quantum convolutional neural
  networks {(QCNN)},'' in \emph{Proceedings of the IEEE International
  Conference on Information and Communication Technology Convergence (ICTC)},
  2020, pp. 236--239.

\bibitem{icoin21oh}
S.~Oh, J.~Choi, J.-K. Kim, and J.~Kim, ``Quantum convolutional neural network
  for resource-efficient image classification: A quantum random access memory
  {(QRAM)} approach,'' in \emph{Proceedings of the IEEE International
  Conference on Information Networking (ICOIN)}, 2021, pp. 50--52.

\bibitem{icoin21choi}
J.~Choi, S.~Oh, and J.~Kim, ``A tutorial on quantum graph recurrent neural
  network {(QGRNN)},'' in \emph{Proceedings of the IEEE International
  Conference on Information Networking (ICOIN)}, 2021, pp. 46--49.

\bibitem{broughton2020tensorflow}
M.~Broughton, G.~Verdon, T.~McCourt, A.~J. Martinez, J.~H. Yoo, S.~V. Isakov,
  P.~Massey, M.~Y. Niu, R.~Halavati, E.~Peters, M.~Leib, A.~Skolik, M.~Streif,
  D.~V. Dollen, J.~R. McClean, S.~Boixo, D.~Bacon, A.~K. Ho, H.~Neven, and
  M.~Mohseni, ``Tensorflow quantum: A software framework for quantum machine
  learning,'' \emph{arxiv}, 2020.

\bibitem{Qiskit}
M.~S.~A. et~al., ``Qiskit: An open-source framework for quantum computing,''
  \emph{arxiv}, 2021.

\bibitem{bergholm2020pennylane}
V.~Bergholm, J.~Izaac, M.~Schuld, C.~Gogolin, M.~S. Alam, S.~Ahmed, J.~M.
  Arrazola, C.~Blank, A.~Delgado, S.~Jahangiri, K.~McKiernan, J.~J. Meyer,
  Z.~Niu, A.~Száva, and N.~Killoran, ``Pennylane: Automatic differentiation of
  hybrid quantum-classical computations,'' \emph{arxiv}, 2020.

\bibitem{mnih2013playing}
V.~Mnih, K.~Kavukcuoglu, D.~Silver, A.~Graves, I.~Antonoglou, D.~Wierstra, and
  M.~Riedmiller, ``Playing {Atari} with deep reinforcement learning,''
  \emph{arXiv:1312.5602}, 2013.

\bibitem{schulman2017proximal}
J.~Schulman, F.~Wolski, P.~Dhariwal, A.~Radford, and O.~Klimov, ``Proximal
  policy optimization algorithms,'' \emph{arXiv:1707.06347}, 2017.

\bibitem{biamonte2021universal}
J.~Biamonte, ``Universal variational quantum computation,'' \emph{Physical
  Review A}, vol. 103, no.~3, p. L030401, 2021.

\bibitem{wiebe2014quantum}
N.~Wiebe, A.~Kapoor, and K.~M. Svore, ``Quantum deep learning,'' \emph{arXiv
  preprint arXiv:1412.3489}, 2014.

\bibitem{chen2020variational}
S.~Y.-C. Chen, C.-H.~H. Yang, J.~Qi, P.-Y. Chen, X.~Ma, and H.-S. Goan,
  ``Variational quantum circuits for deep reinforcement learning,'' \emph{IEEE
  Access}, vol.~8, pp. 141\,007--141\,024, 2020.

\bibitem{jerbi2021variational}
S.~Jerbi, C.~Gyurik, S.~Marshall, H.~J. Briegel, and V.~Dunjko, ``Variational
  quantum policies for reinforcement learning,'' \emph{arXiv preprint
  arXiv:2103.05577}, 2021.

\bibitem{NEURIPS2019_9015}
A.~e.~a. Paszke, ``Pytorch: An imperative style, high-performance deep learning
  library,'' in \emph{Advances in Neural Information Processing Systems
  (NIPS)}, 2019, pp. 8024--8035.

\bibitem{1606.01540}
G.~Brockman, V.~Cheung, L.~Pettersson, J.~Schneider, J.~Schulman, J.~Tang, and
  W.~Zaremba, ``Openai gym,'' 2016.

\end{thebibliography}
\end{document}